# Enhanced Precision in Rainfall Forecasting for Mumbai: Utilizing Physics-Informed ConvLSTM2D Models for Finer Spatial and Temporal Resolution

Ajay Devda[1], Akshay Sunil[1], Murthy R[3], B Deepthi[1]

1. Interdisciplinary Programme in Climate Studies, Indian Institute of Technology, Bombay
2. Department of Civil Engineering, Indian Institute of Technology, Bombay
3. Department of Civil Engineering, BMS college of Engineering

[*]Corresponding author: E-mail: akshaysunil172@gmail.com

## Abstract

Forecasting rainfall in tropical areas is challenging due to complex atmospheric behavior, elevated humidity levels, and the common presence of convective rain events. In the Indian context, the difficulty is further exacerbated because of the monsoon intra-seasonal oscillations, which introduce significant variability in rainfall patterns over short periods. Earlier investigations into rainfall prediction leveraged numerical weather prediction methods, along with statistical and deep learning approaches. This study introduces a nuanced approach by deploying a deep learning spatial model aimed at enhancing rainfall prediction accuracy on a finer scale. In this study, we hypothesize that integrating physical understanding improves the precipitation prediction skill of deep learning models with high precision for finer spatial scales, such as cities. To test this hypothesis, we introduce a physics-informed ConvLSTM2D (Convolutional Long Short-Term Memory 2D) model to predict precipitation 6hr and 12hr ahead for Mumbai, India. We utilize ERA-5 reanalysis data with hourly time steps spanning from 2011 to 2022 to select predictor variables, including temperature, potential vorticity, and humidity, across various geopotential levels. The ConvLSTM2D model was trained on the target variable precipitation for 4 different grids representing different spatial grid locations of Mumbai. The Nash-Sutcliffe Efficiency (NSE), utilized to evaluate the precision of 6 and 12 hours ahead precipitation forecasts, yielded ranges of 0.61 to 0.68 for 6-hour predictions and 0.58 to 0.66 for 12-hour predictions during the training phase. In the testing phase, the NSE values range from 0.42 to 0.51 for 6-hour forecasts and from 0.47 to 0.58 for 12-hour forecasts, respectively. These values highlight the model's high accuracy and its capacity to capture variations. Thus, the use of the ConvLSTM2D model for rainfall prediction, utilizing physics-informed data from specific grids with limited spatial information, reflects current advancements in meteorological research that emphasize both efficiency and localized precision.

## 1. Introduction

Precipitation particularly rainfall acts as a critical determinant in the initiation and development of natural hazards such as floods, which hugely disrupt human and natural processes. Forecasting precipitation at high resolution hours in advance meets the socioeconomic demands of various sectors that rely on weather information to make vital decisions (Ravuri et al., 2021). Coastal cities are particularly vulnerable to rainfall due to their high economic activities, transportation needs, dense infrastructure and urban sprawl particularly vulnerable to rainfall due to their high economic activities, transportation needs, dense infrastructure, and urban sprawl (Snow et al., 2012; Gill and Malamud, 2014). Rainfall intensity highly varies across small spatial and temporal scales, which creates a necessity to forecast rainfall for finer grids reference to a locality or region within a few kilometres' radius for cities(Liu and Niyogi, 2019; Moron and Robertson, 2020). Short term, localized rainfall forecasting is important for applications like flash flood warnings, management of dams and reservoirs, and transportation planning (Ali et al., 2014; Yadav and Ganguly, 2020). The temporal and spatial fluctuations of rainfall are significantly impacted by various factors including atmospheric drivers (such as atmospheric circulation), temperature, wind patterns, terrain, and humidity (Kishtawal et al., 2010; Zhou et al., 2019). These factors govern rainfall patterns, leading to variations in rainfall over short distances and time periods. The chaotic behaviour of atmosphere makes it extremely sensitive to small changes in dynamics leading to vastly different weather phenomenon (Lorenz, 1963). All these complex physical systems along with anthropogenic attributions which cause climate change, make rainfall a complex phenomenon to predict (Trenberth et al., 2003; Allan and Soden, 2008). Considering the dynamics of the Earth system, alongside local atmospheric flows and meteorological traits, aids in accurately depicting the local-scale phenomenon for rainfall predictions (Pielke et al., 1992; Trenberth and Asrar, 2014).

Previous studies have examined the connections between rainfall rates on an hourly basis and different atmospheric variables to understand the choice of predictor variables for precipitation forecast (Lepore et al., 2016; Mitovski and Folkins, 2014). Lepore et al. (2016) studied how hourly rainfall correlates with humidity and CAPE (Convective Available Potential Energy) across the United States from 1979 to 2012. This study concluded that moisture availability and vertical instability significantly affect rainfall occurrence. Mitovski and Folkins (2014) analysed high rainfall events across four regions: Southeast China, Tropical Brazil, Western Tropical Pacific, and Southeast United States. They utilized 13 years of rainfall data from the Tropical Rainfall Measuring Mission to investigate high precipitation events in these diverse regions. This study concluded that atmospheric dynamics, including mass divergence, potential vorticity, and relative vorticity, are interconnected and influence convective activity, which in turn can result in rainfall. Temperature, humidity and vorticity at different geopotential heights affect the cloud formation and considered as potential predictor for rainfall forecasting (Bansod, 2005).

Previous meteorological studies have utilized various methodologies, such as numerical weather modeling (Warner, 2011; Holton and Hakim, 2012), statistical methods (Wilks, 2011; Fischer et al., 2012), and deep learning techniques (Shi et al., 2015.; Sønderby et al., 2020; Yadav and Ganguly, 2020; Castro et al., 2021; Espeholt et al., 2022) to enhance precipitation predictions across a range of temporal scales from hourly to annual. However, numerical weather prediction models face significant challenges in tropical regions due to the inherent complexity and variability of weather patterns. For instance, the El Niño Southern Oscillation (ENSO) is a prominent example of such a factor. Also, there are other influential phenomena, such as the Indian Ocean Dipole (IOD) and the Madden-Julian Oscillation (MJO), which can also impact the distribution and intensity of monsoonal rains (Ray et al., 2022).

This calls for a need to employ modern computing capabilities to enhance the prediction skill of various climate variables (Castro et al., 2021). Recent advancements in computational capabilities such as GPUs and parallel computing, have significantly enhanced the feasibility of employing deep learning techniques in atmospheric and climate sciences research (Lecun et al., 1998; Sønderby et al., 2020; Yadav and Ganguly, 2020). Developing deep neural network architectures for atmospheric systems in particular regions is a challenging endeavor, requiring an in-depth comprehension of atmospheric physics and extensive experimentation (Kreuzer et al., 2020; Tong et al., 2022). The complexity is further amplified by the numerous hyperparameters within the network(Xiao et al., 2019; Sønderby et al., 2020; Espeholt et al., 2022; Dehghani et al., 2023), requiring a delicate balance between theoretical knowledge and empirical testing to achieve accurate atmospheric modelling. Prior studies have shown that hybrid models that blend deep learning with physical science enable more precise local rainfall predictions (Castro et al., 2021; Gao et al., 2021). The convolution neural networks have been combined with long short-term memory (LSTM) to predict rainfall in spatial and temporal dimensions (Castro et al., 2021; Khan and Maity, 2020; Yadav and Ganguly, 2020; Castro et al., 2021; Li et al., 2022; Dehghani et al., 2023; Li et al., 2023). For instance, Castro et al. (2021) employed a new deep-learning framework called STConvS2S to capture spatiotemporal predictive patterns by utilizing meteorological variables, aiming to enhance the accuracy of precipitation forecasting.

Convolutional LSTM networks combine the advantages of Convolutional Neural Networks (CNNs) and Recurrent Neural Networks (RNNs), enabling them to capture both spatial and temporal information effectively. This integration enhances the accuracy of weather prediction compared to conventional machine learning models, as demonstrated in various studies (Nastos et al., 2014). Gao et al. (2021) reviewed different deep-neural network methods for short-term rainfall prediction and found that ConvLSTM outperforms other methods. Several attempts have been made to utilize the ConvLSTM to predict rainfall (Shi et al., 2017; Kumar et al., 2020; Yadav and Ganguly, 2020; Kumar et al., 2021). For instance, Yadav and Ganguly (2020) highlight the challenges in predicting very short-term distributed quantitative precipitation. In this study, the authors explore suitable architecture of the ConvLSTM model to enhance short-term rainfall forecasting, employing data from NASA's North American Land Data Assimilation System (NLDAS) across the United States. Yasuno et al. (2021) predicted 6-hours ahead rainfall using ConvLSTM for Japan. They utilized 37000 hourly radar images points from 2006 to 2019 for training and validating the model. The model attains high levels of accuracy in RMSE and MAE, demonstrating ConvLSTM's effectiveness in predicting rainfall, especially at finer spatial resolutions. Therefore, ConvLSTM could be suitable for identifying the abrupt alterations in the precipitation field provided it has encountered similar patterns during training. Sønderby et al. (2020) introduced 'MetNet', a neural weather model to forecast precipitation 8 hours ahead using radar and satellite data for the USA. Espeholt et al. (2022) presented 'MentNet-2' a physics-based deep-learning model to predict 12-hour rainfall at 1 km spatial resolution for the USA. The model takes temperature, humidity, wind speed and direction, pressure, and satellite imagery as predictor variables. In the Indian context, Khan and Maity (2020) developed a hybrid Deep Learning (DL) model by combining the 1 D CNN (One dimensional Convolutional Neural Network) with a Multi-Layer Perceptron (MLP) for predicting daily rainfall up to 5 days ahead. Nine meteorological variables related to precipitation were used as predictors for twelve locations in Maharashtra in this study.

Mumbai, primarily situated along the Arabian coast, is often susceptible to urban floods and socioeconomic losses. Western Ghats, the barriers at western coast, arrest the southwest monsoon rainfall and bring intensified rainfall to the city making it difficult to accurately predict Mumbai's rainfall (Mohanty et al., 2023). Numerous investigations have examined the rainfall patterns and extremity in Mumbai. For example, Gope et al. (2016) presented a Stacked Auto-Encoder (SAE) based deep-learning model to forecast heavy rainfall in Mumbai and Kolkata up to 6 to 48 hours ahead. They have used climatic variables such as temperature and relative humidity, as well as atmospheric variables such as northward wind and eastward wind, for rainfall prediction. However, they did not account for extreme events, and the spatial variability was not adequately captured. Previous studies have highlighted the need for exploring and integrating atmospheric variables in deep-learning models to improve precipitation forecasts across different geographical locations and climatic conditions (Castro et al., 2021). Building upon this, in the present study, we hypothesize that the physics-informed deep learning, through the integration of carefully selected atmospheric variables can improve the rainfall forecast skills. We test this hypothesis for the Coastal City Mumbai, a region with high variability in rainfall. Mumbai receives a significant portion of its rainfall during southwest monsoon due to the significant seasonal changes in wind patterns. Mohanty et al. (2023) identified three systems responsible for extreme rainfall behaviors in Mumbai: the offshore trough, mid-tropospheric cyclones, and Bay of Bengal (BoB) Depression. This highlights that atmospheric variables are the primary influencers of the city's precipitation patterns. To date, there have been no studies testing the efficiency of physics-informed deep learning methods for forecasting rainfall in finer scales (like urban cities), particularly in Mumbai, using minimal atmospheric variables. In the present study, the selection of suitable atmospheric variables (predictors) to train the deep-learning based ConvLSTM2D model is guided by the understanding of the underlying atmospheric processes that cause precipitation (target). The ConvLSTM2D model can learn from the spatio-temporal patterns of the predictors to predict the target variable. The model highlights the significance of physics-based variables that influence rainfall for deep learning methods. This combination can be further utilized to offer precise forecasts of rainfall events at more detailed spatial scales in Mumbai and other similar regions, aiding in mitigating losses and enhancing micro-level planning and actions by relevant agencies.

The rest of this manuscript is organized as follows: Details of the study area and the data used for the analysis are provided in Section 2. Section 3 outlines the methodology, where we describe the architecture of the ConvLSTM2D model. The findings and analysis appear in Section 4, while the study's conclusions are detailed in Section 5.

## 2. Study Area and Data
### 2.1 Study Area

Located in the Western Ghats region, Mumbai falls within the tropical monsoon climate zone. Its geographical coordinates range from 18° to 19.2° N and 72° to 73° E, encompassing a total area of 437.79 km² (Figure 1). The city receives an average annual precipitation of 2450 mm Singh et al., 2017; Mohanty et al., 2023;). Mumbai experienced an unexpected 944mm of rainfall in a single day in the year 2005 (Singh et al., 2017) which cause high lives and economic loss. Such high rainfall extremes in a short period are the main cause of urban floods; a timely and accurate projection of rainfall gives planning agencies enough time to prevent waterlogging and save human and economic losses. Hence, there arises a critical need for a robust nowcasting system in coastal cities such as Mumbai. This system would provide precise information regarding the location and intensity of rainfall, ensuring that both individuals and the public disaster response system are well-prepared.

Mohanty et al. (2023) emphasized that Mumbai's extreme rainfall events (ERF), such as the 944 mm downpour on 26 July 2005, are influenced by three main rain-bearing systems: offshore troughs, mid-tropospheric cyclones (MTC), and Bay of Bengal depressions. These mechanisms play a crucial role in determining the city's precipitation trends, resulting in Extreme Rainfall

Events (ERF) surpassing 204.5 mm within a single day, occurring roughly biennially throughout the summer monsoon period (Zope and Eldho, 2012; Gope et al., 2016; Singh et al., 2017). This connection underscores the significant impact of these atmospheric phenomena on the frequency and intensity of Mumbai's monsoon extremes. Due to its geographical and physical characteristics, the city faces numerous risks, including cyclones, floods, earthquakes, and landslides, triggered by both natural events and human activities. Due to high and unplanned urbanization, low laying areas and conjunction in the drainage system, during monsoons almost every year city faces water logging which affects transportation and sometimes leads to loss of lives and economic damage.

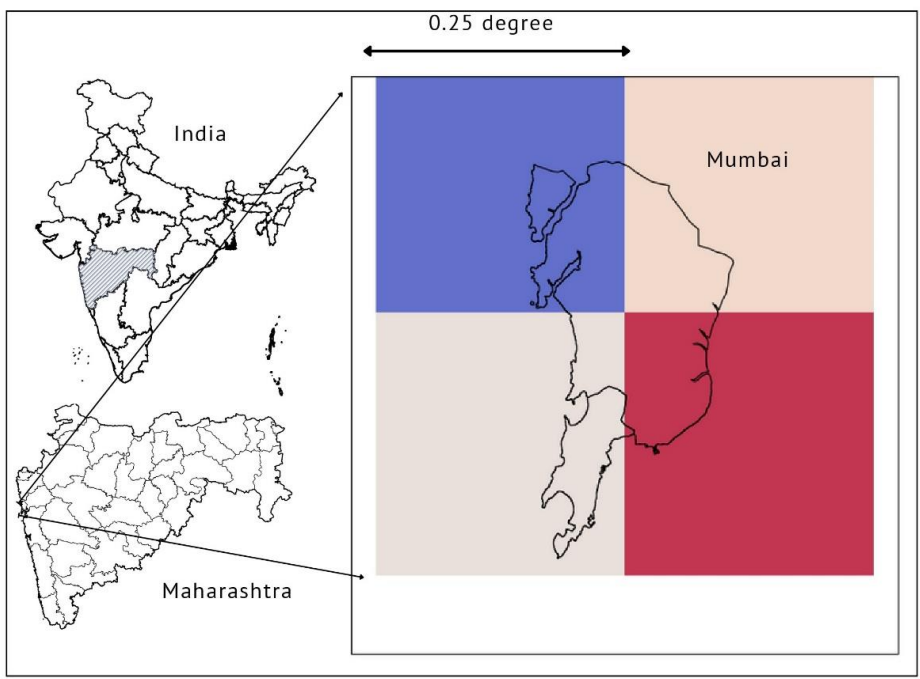

**Figure 1.** Location of Mumbai and the grids covered by ERA5 reanalysis products

**2.2 Data**

**2.2.1 Identifying Physics-Informed Variables for High-Resolution Spatial Forecasting of Near-Term Precipitation**

Developing a deep learning model informed by physics for accurate precipitation prediction requires a comprehensive grasp of atmospheric dynamics, including the processes of evaporation, formation and condensation of clouds, atmospheric movement, and the complex interactions between the atmosphere's various layers. This understanding helps in selecting the most suitable variables as predictors for the deep learning model (Kashinath et al., 2021; Shen and Lawson, 2021; Teufel et al., 2023). The variable selections have been done by a deep literature review of studies related to rainfall predictions and deep learning methods. O'German and Schneider (2009) emphasize the significance of taking into account changes in both specific humidity and temperature for precise prediction of precipitation extremes. These variables directly influence cloud condensation processes and the overall atmospheric stability, thereby causing variations in precipitation intensity and distribution. Weyn et al. (2020) investigated the capability of CNNs in forecasting weather patterns, focusing on the 500-hPa geopotential height as a key input variable. Their research illustrates that CNNs, trained on historical weather data, not only surpass traditional forecast benchmarks but also deliver reliable weather forecasts (500-hPa geopotential height) up to 14 days ahead. Studies have also compared numerical weather prediction (NWP), with deep convolutional neural network (CNN) algorithm finding that CNNs can significantly enhance forecasting accuracy for severe convective weather events (Zhou et al., 2019). The present study focused on developing a deep learning model that combines convolutional layers for spatial analysis and LSTM for temporal analysis to predict rainfall variability and timing (e.g., 6, and 12 hours ahead) at finer spatial scales, such as cities. This model aims to aid meteorological research and assist water resource managers, disaster response teams, and decision-makers in metropolitan areas like Mumbai. It also seeks to provide valuable insights for the urban community and farmers, contributing to a decision support system for rainfall nowcasting. Historically, a range of indicators have been employed to enhance precipitation forecasts through downscaling, including sea level pressure (Cavazos, 1999), geopotential height (Kidson and Thompson, 1998), geostrophic vorticity (Wilby and Wigley, 2000), and wind velocity (Murphy, 1999).The choice of predictors varies depending on factors like regional distinctions, characteristics of large-scale atmospheric circulation, seasonal variations, and geomorphological attributes (Anandhi et al., 2009).

### 2.2.2 Data used

Rainfall is highly sensitive to atmospheric processes such as temperature gradients, vapour and cloud formation and movement (Lenderink and Fowler, 2017). The availability of ERA5 reanalysis data through ECMWF (European Centre for Medium-Range Weather Forecasts) provides the most suitable variables, leading to improved representation of spatial and temporal diversity with higher accuracy, particularly when attributing rainfall. The atmospheric product of the ERA reanalysis is prepared by complex data assimilation and reconstruction of satellite products and ground observation from weather stations and radars (Hersbach et al., 2020). The dataset provides hourly information on a variety of meteorological variables at 0.25° resolution, on a global scale covering the period from 1979 to the present (Taszarek et al., 2020; Jiao et al., 2021). Research has underscored enhancements in the ERA5 dataset's spatial and temporal resolution over its predecessor, ERA-Interim, establishing it as a crucial asset for predictive modelling and analytical tasks (Nogueira, 2020; Jiao et al., 2021). Hence, ERA5 reanalysis datasets have found extensive applications in climate change research (Letson et al., 2021). Previous studies have also highlighted that ERA-5's precipitation data demonstrates notable accuracy and outperforms other reanalysis products for India (Mahto and Mishra, 2019).

In this study, we employ data from the ERA 5 reanalysis, selecting 11 predictor variables for forecasting precipitation. The predictor variables selected for this study consist of temperature (t), relative humidity (rh), and potential vorticity at 250 hPa, 500 hPa, and 850 hPa. The remaining three predictors are total cloud coverage observed (tcc), high-level cloud coverage observed (hcc), and atmospheric pressure at the surface level (sp). Descriptions, units, and references for each of these predictors are provided in Table 1. Hourly records of these physics-informed variables were downloaded and compiled for the duration spanning from 2011 to 2022, providing a substantial dataset for analysis, model training, prediction generation, and subsequent validation. For the ConvLSTM2D model, the target variable chosen is "total precipitation (tp)," encompassing both convective and widespread rainfall events (Terblanche et al., 2022). Previous studies have used ERA-5 "tp" as a measure of rainfall (Jiang et al., 2023).

**Table 1:** Details of the 11 predictors considered in the study.

| Variable | Description | Unit | Reference |
|---|---|---|---|
| t (250 hpa) | Temperature at the geopotential height of 250 hPa | K | |
| t (500 hpa) | Temperature at the geopotential height of 500 hPa | K | (Khan and Maity, 2020) |
| t (850 hpa) | Temperature at the geopotential height of 850 hPa | K | |
| rh (250 hpa) | Relative humidity at the geopotential height of 250 hPa | % | (Khan and Maity, 2020; Salaeh et al., 2022) |
| rh (500 hpa) | Relative humidity at the geopotential height of 500 hPa | % | |
| rh (850 hpa) | Relative humidity at the geopotential height of 850 hPa | % | |
| pv (500 hpa) | Potential vorticity at the geopotential height of 500 hPa | K m$^2$ kg$^{-1}$ s$^{-1}$ | |
| pv (850 hpa) | Potential vorticity at the geopotential height of 850 hPa | K m$^2$ kg$^{-1}$ s$^{-1}$ | |
| tcc | Total cloud coverage observed | % | |
| hcc | High-level cloud coverage observed | % | |
| sp | Atmospheric pressure at surface level | Pa | (Khan and Maity, 2020) |

## 2.2.3. Data preparation

The study area consists of four grids covering the spatial extent of Mumbai. The global ERA-5 reanalysis dataset was downloaded from the 'ERA-5 hourly data on single levels from 1940 to present' (Hersbach et al., 2020). The data was further cropped using the latitude and longitude extents of Mumbai for the 12 years for the period 2011-2022. Consequently, our dataset comprises 105192 hourly observations of chosen variables for the purpose of training and evaluating the ConvLSTM2D model. We further normalise the data to the range [0,1] to scale the units of different atmospheric variables.

## 3. Methodology

### 3.1 Model Architecture

In this study, we employ the ConvLSTM2D network, which is an extension of the traditional LSTM and utilises convolution operations in both input-to-state and state-to-state transitions to handle spatiotemporal data more effectively. This integration allows the model to capture spatial dependencies alongside temporal ones, making it particularly suited for tasks like precipitation nowcasting. Unlike standard LSTM layers that process data one point at a time, ConvLSTM layers use convolution operations within the LSTM cell, making them suitable for spatial data with temporal dependencies. The model adapted Hierarchical Feature Learning where the first layer captures wide range of features and second layer with fewer filters refine these features and focus on most relevant data.

The methodological flow chart illustrating the steps involved, such as input variable selection for the 2-dimensional spatial grids, data preparation and normalization, the ConvLSTM2D model layers, and the model output for the selected grids, is shown in Figure 2. Now, the ConvLSTM model architecture is discussed (Figure 3). The ConvLSTM controls the data flow inside the cell through the forget gate ($F_t$), input gate ($I_t$), and output gate ($O_t$). The amount of information that should be forgotten and retained by the model is determined by forget gate. At the end of each iteration, forget gate devices to discard or transmit the relevant information. Memory cell ($C_{t-1}$) in the ConvLSTM2D models acts like an accumulator of information at every state thereby aiding information accumulation. Several self-parametrized controlling gates are employed by the LSTM to access the cell. Input gate activation accumulates information to cell. Therefore, the input gate opens the way to integrate the new data in cell and add information to the long-term memory. The forget gate aids in forgetting information from the past cell status ($C_{t-1}$). Finally, the output gate propagates the updated information to the next LSTM cell. Here, the values in the output gate is multiplied with the updated cell information by passing the cell state after updating it through a activation function (tanh) to calculate the hidden state ($H_t$).

Mathematical expression of the ConvLSTM layer for each time step *t* is as follows:
1. Input gate

$$I_t = \sigma(W_{xi} * X_t + W_{hi} * H_{t-1} + W_{ci} \odot C_{t-1} + b_i)$$

2. Forget gate

$$F_t = \sigma(W_{xf} * X_t + W_{hf} * H_{t-1} + W_{cf} \odot C_{t-1} + b_f)$$

3. Cell state

$$Ct_t = F_i\, C_{t-1} + I_t \odot \tanh(W_{xc} * H_{t-1} * + W_{hc} * H_{t-1} + b_c)$$

4. Output gate

$$O_t = \sigma(\boldsymbol{W_{zo}} * X_t + W_{ho} * H_{t-1} + W_{co} \odot C_t + b_o)$$

5. Hidden state

$$H_t = O_t \odot \tanh(C_t)$$

where * denotes the convolution operation and ⊙ is elementwise Hadmard product. The sigmoid activation function and state's weighed connections are represented by σ and W, respectively (Shi et al., 2017; Kumar et al., 2020; Moishin et al., 2021). σ ensures the gate activations values ranges from 0 to 1, and tanh is a hyperbolic tangent activation function that introduces nonlinearity on the cell update in the operations, and values range from -1 to 1. The term $W_{xc}, W_{xf}, W_{xi}, W_{xo}$ are the weights for the input ($X_t$) to different gates such as cell state, forget gate, input gate, and output gate. $W_{hc}, W_{hi}, W_{ho}, W_{hf}$ are the weights for the hidden state to cell state, input gate, output gate, and forget gate. $b_c, b_i, b_f, b_o$ are bias terms for the cell state, input gate, forget gate, and output gate. $H_t$ and $C_t$ are the hidden and cell state for the current timestep, and $X_t$ is input at the current timestep. $C_{(t-1)}$ and $H_{(t-1)}$ are the cell and hidden states for previous timesteps.

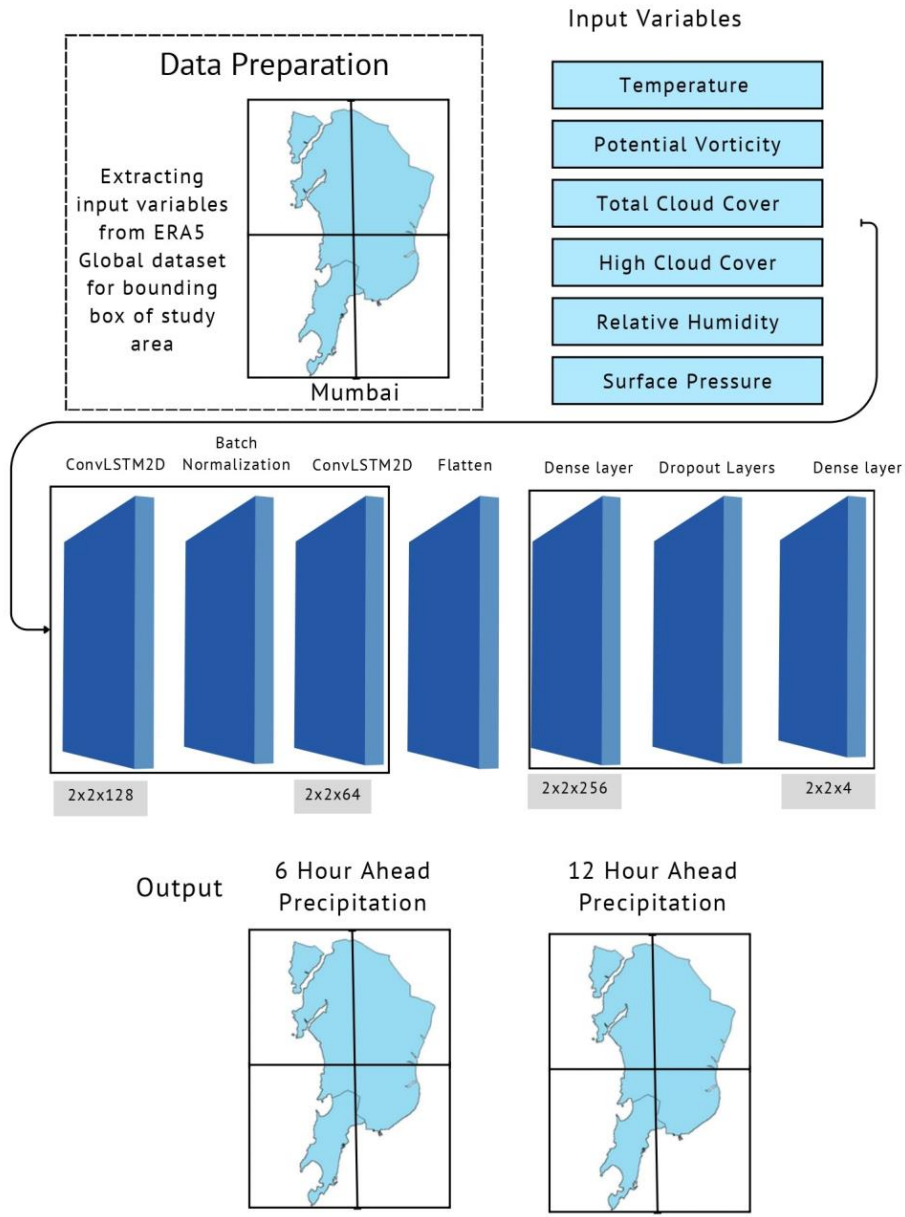

**Figure 2.** Methodological flow chart for the ConvLSTM2D model

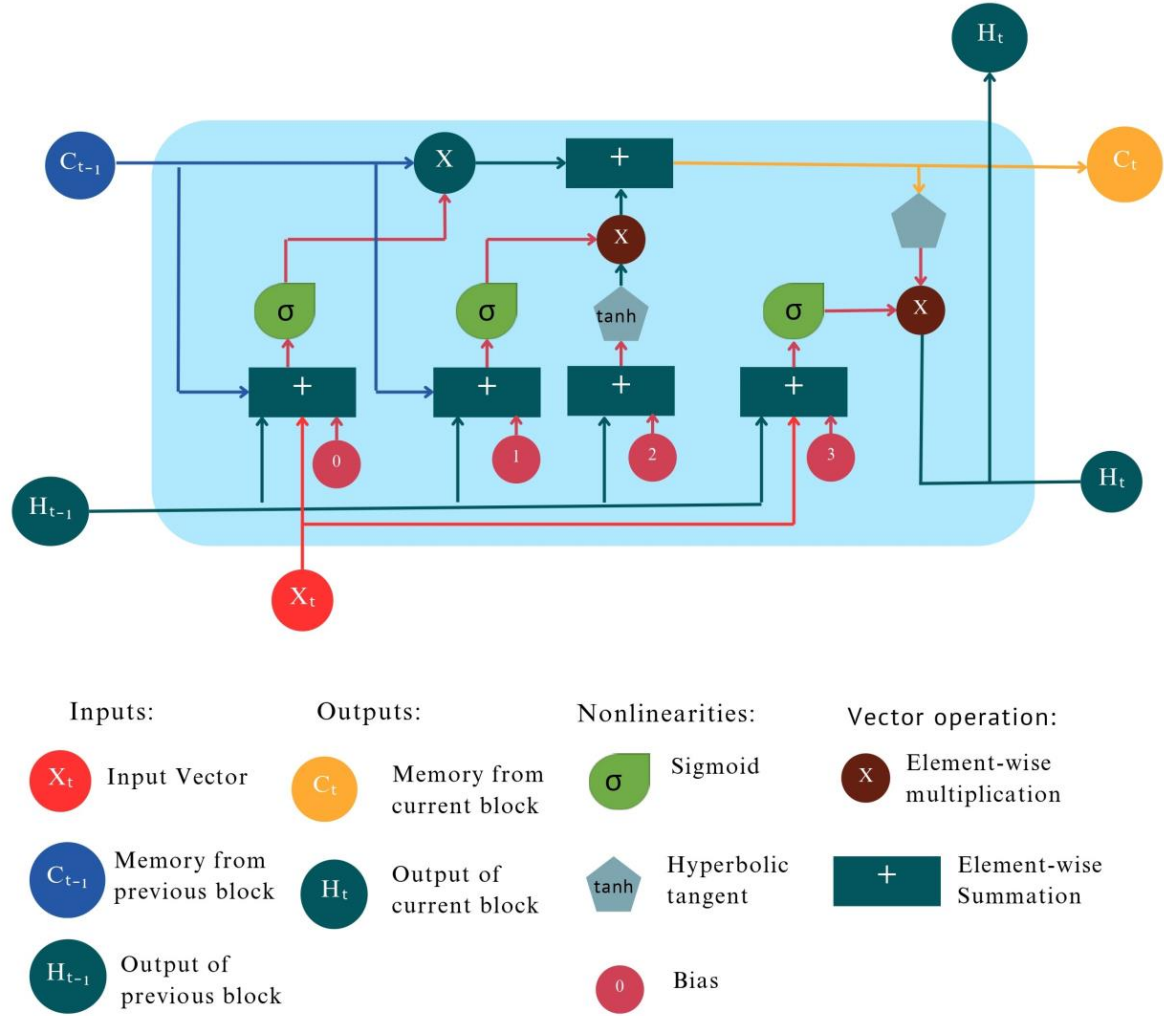

**Figure 3.** Illustration of a single Long Short-Term Memory (LSTM) unit architecture. Each unit of LSTM composes a cell ($C_t$), an input gate ($I_t$), an output gate ($O_t$), and a forget gate ($F_t$), all of which work together to regulate the flow of information.

The present model used in the study consists of two Conv LSTM layers with different configurations. In a ConvLSTM2D model, the filters extract different features from the input data and captures various aspects from the spatial information (Ravuri et al., 2021; Guastavino et al., 2022; Bi et al., 2023), and therefore a greater number of filters implies model learns from a rich representation of data. The first layer starts the process of extracting relevant features across both dimensions, and the second layer consolidates these features, focusing on the most relevant temporal aspects. In the present model, the initial ConvLSTM Layer has 128 filters. The kernel size of this Layer is (2, 2). The second ConvLSTM Layer has 64 filters, fewer than the first layer, which helps in progressively reducing the feature space dimensionality, making the model computationally efficient. 'Relu (Rectified Linear Units) ' activation function is employed in the model because of its computational simplicity and introduces non-linearity to the model, which allows the model to learn from complex patterns. The input shape parameter defines the shape of the input data, which includes the sequence length and spatial dimensions. Finally, the return sequence set to false so the layer will output only to final results to sequins processing, which is suitable for making predictions. The hyperparameters of the ConvLSTM2D model in provided in Table S1.

## 3.2 Training and prediction

The present ConvLSTM2D employs a prediction method to generate outputs for training, and testing data, leveraging the learned weights to estimate rainfall patterns from the input features. For the ConvLSTM2D deep learning model, 85% of the dataset, comprising 105192 hours of predictor and target variables, is allocated to training, with the remaining 15% designated for testing. From the testing dataset, a 15% subset is further designated for model validation. The ConvLSTM2D model processes the dataset initially structured as a 2x2 spatial grid by reshaping it (utilizing reshape(-1, 2, 2)) for compatibility with its training architecture ensuring that the spatial dimensions of the model's outputs align with input grid dimensions. Subsequently, the model's performance is evaluated by comparing predictions with the actual data using Correlation Coefficient (CC), Nash-Sutcliffe Efficiency (NSE), and Normalized Root Mean Square Error (NRMSE).

## 4. Results and Discussion

Initially, the correlation among the predictors and between the predictors and rainfall is assessed to evaluate the relevance of the chosen variables as predictors in the ConvLSTM2D model. The details regarding the selection of these predictors are already outlined in Section 2.2.1. Subsequently, these variables are utilized to forecast rainfall during both the training and testing phases at 6-hour and 12-hour time intervals ahead. The results for the 6-hour and 12-hour time intervals are presented separately.

### 4.1. Correlation Matrix

The correlation matrix, displaying the interrelationships among the predictors and between the predictors and rainfall, is presented in Figure 4. As observed in Figure 4, total precipitation (tp) exhibits the highest correlation (0.43) with relative humidity at 500 hPa (rh 500 hPa), while the minimum correlation (-0.36) is obtained for surface pressure. In addition to rh 500 hPa, predictors showing positive correlation with 'tp' include total cloud cover (tcc), high cloud cover (hcc), relative humidity at 250, 500, and 850 hPa (rh 250 hPa, rh 500 hPa, and rh 850 hPa), potential vorticity at 500 and 850 hPa (pv 500hPa, and pv 850 hPa), and temperature at 250 and 500 hPa (t 250 hPa, and t 500 hPa). Both total cloud cover (tcc) and high cloud cover (hcc) exhibit the same positive correlation value (0.34) with total precipitation, indicating that an increase in cloud cover leads to increased precipitation. This finding aligns with previous studies in atmospheric science that have investigated the relationships between cloud cover and precipitation extremes (Mendoza et al., 2021; Zhong et al., 2021). For instance, the presence of high clouds often signifies the approach of a front or a low-pressure system, both of which are associated with increased precipitation

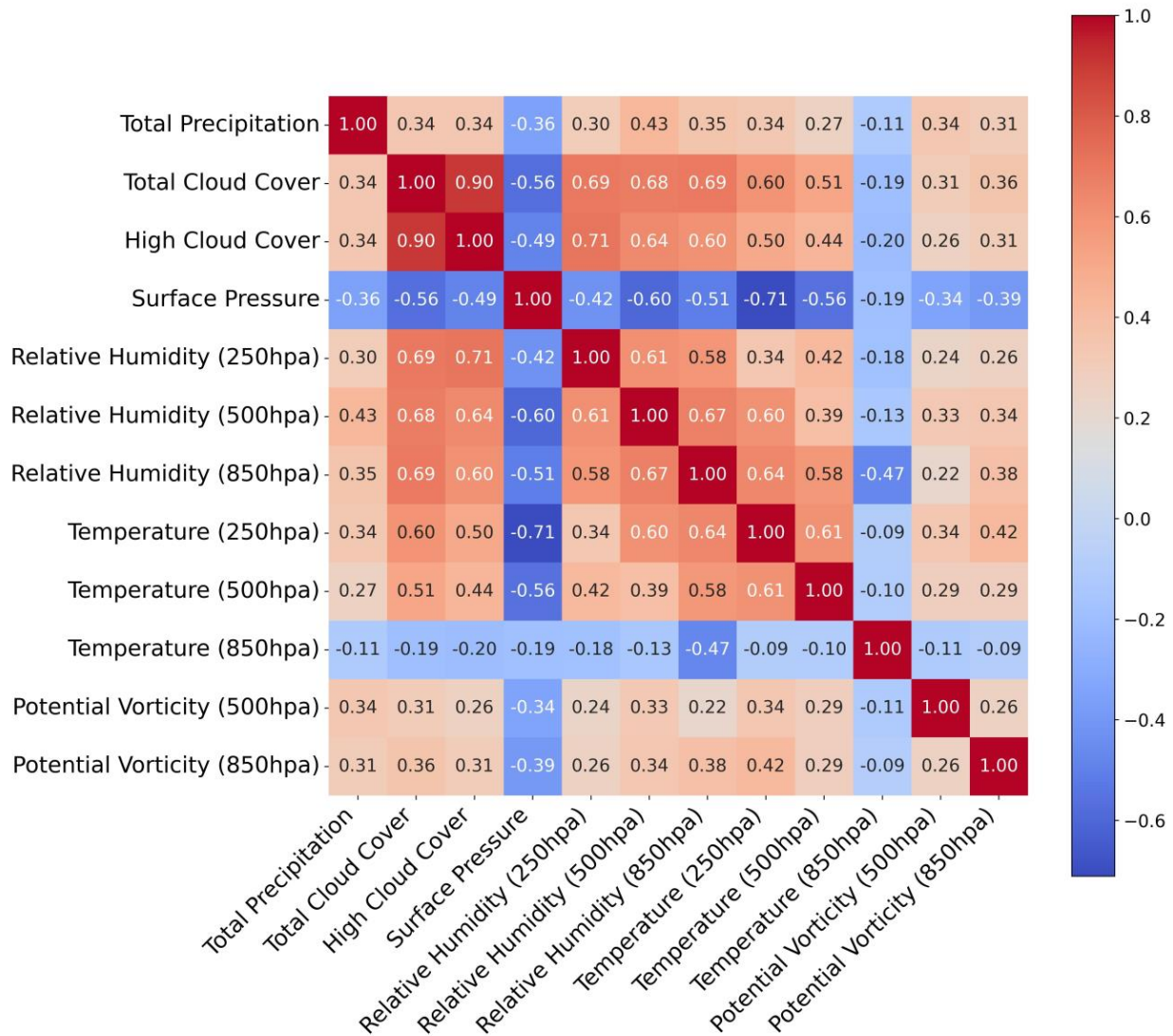

**Figure 4.** Heatmap displays the correlation matrix of atmospheric variables, ranging from -0.71 to 1. Positive correlations are depicted in red, while negative correlations are shown in blue. A correlation value of 1, indicating the correlation of a variable with itself, is represented by a dark red color in the heatmap.

Relative humidity at various tropospheric levels—250 hPa, 500 hPa, and 850 hPa—exhibits positive correlations with total precipitation, with correlation values of 0.3, 0.43, and 0.35 respectively. This suggests an increasing likelihood of precipitation as relative humidity rises, with the mid-tropospheric level of 500 hPa showing the strongest correlation. This finding is consistent with earlier studies emphasizing the significant influence of relative humidity on precipitation probabilities.

Potential Vorticity at 500 hPa and 850 hPa shows a positive correlation with total precipitation of 0.34 and 0.31 respectively, indicating a statistical relationship between higher potential vorticity and temperature. Consistent with earlier studies, potential vorticity acts as a significant marker for air masses, preserved in adiabatic and friction-free movement, making it a valuable identifier for air masses (Wallace and Hobbs, 2006; Luhunga and Djolov, 2017). High potential vorticity is associated with low-pressure systems, leading to enhanced precipitation events (Wallace and Hobbs, 2006). However, potential vorticity at 250 hPa is not considered as a predictor in this study due to its negligible correlation with total precipitation. This is because potential vorticity, despite being a measure of the rotation and stratification of the atmosphere, requires conjunction with other factors like humidity and temperature gradients to effectively influence precipitation (Holton and Hakim, 2012).

The correlation between total precipitation and temperature varies at different geopotential heights. For instance, the correlation between 'tp' and temperature at 500 hPa is 0.27, suggesting that warmer air in mid-tropospheric levels contributes to the instability required for cloud formation and subsequent precipitation (White et al., 2016). This could be linked to the concept of convective available potential energy (CAPE), where warmer temperatures aloft can lead to a more unstable atmosphere, potentially resulting in thunderstorms and heavy precipitation under the right conditions (Markowski and Richardson, 2014).

Among all the 11 predictors considered, the predictors showing negative correlation with 'tp' are surface pressure and temperature at 850 hPa only. The negative correlation between total precipitation and surface pressure aligns well with established meteorological principles, indicating that lower surface pressures are associated with storm systems that can result in increased precipitation events (Davies-Jones and Markowski, 2013). Moreover, surface pressure affects atmospheric flow patterns, thereby altering the geographic distribution and strength of rainfall. The correlation between total precipitation and temperature at 850 hPa is -0.11.

**4.2 Results for 6-hour and 12-hourTime Intervals**

The ConvLSTM2D model is trained using 11 selected predictors to forecast rainfall at 6-hour and 12-hour time intervals. Time series plots are employed to visually represent the relationship between observed and predicted rainfall values. In Figure 5, the time series plot illustrates this relationship during the training phase, with data points at 6-hour intervals. The observed(actual) data points are represented by a blue color, whereas forecasted data points are illustrated using a red color. Given that the study area comprises four grids, Figure 5 displays the time series plots for each of these grids.

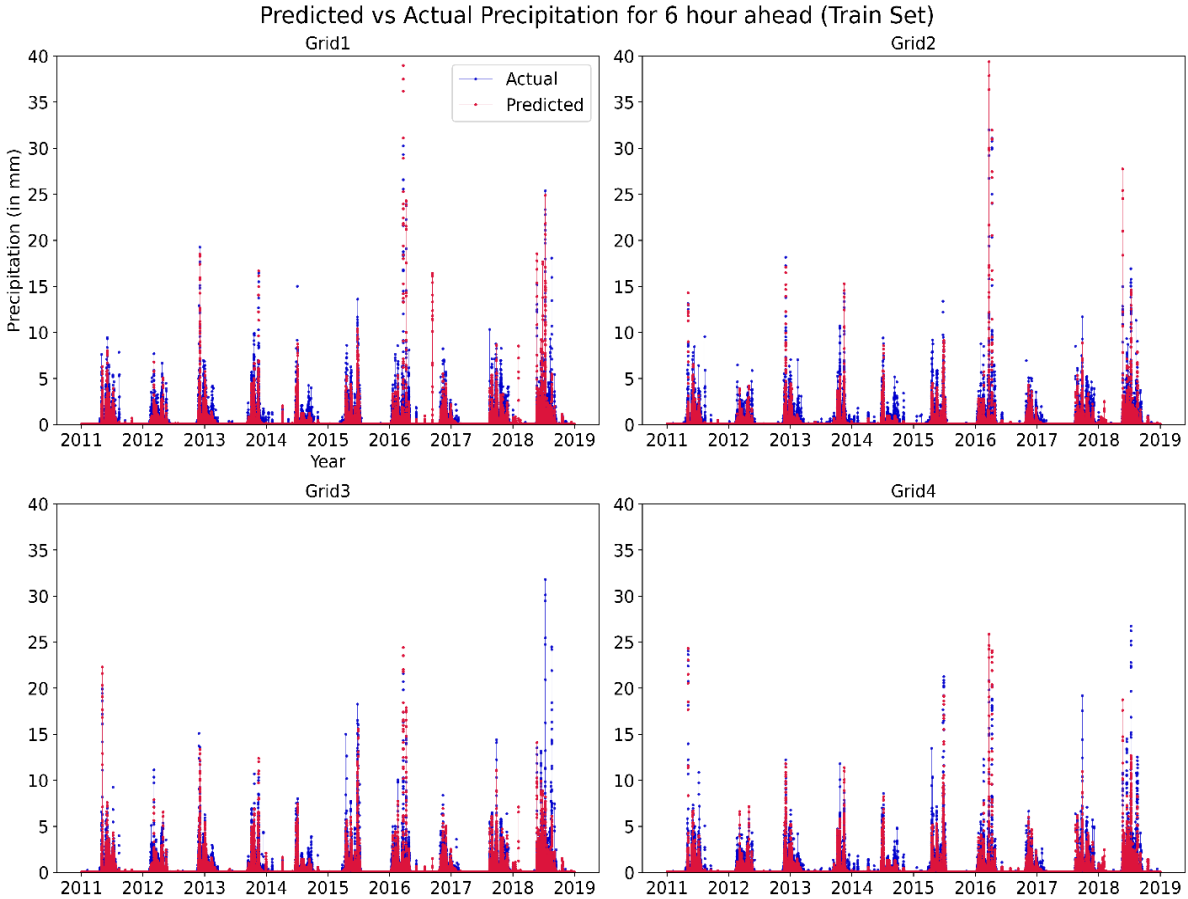

**Figure 5**. Time series comparisons of actual and forecasted rainfall across four grids at 6-hour intervals during the training phase. The observed data points are represented by a blue color, whereas forecasted data points are illustrated using a red color.

As depicted in Figure 5, the predicted rainfall obtained at 6-hour intervals during the training phase shows good agreement with the observed rainfall time series across all four grids. However, there is an overestimation of rainfall at grid 2, while at grid 3, there is an underestimation. For the training data, the actual maximum rainfall values for the four grids range from 26.74 mm to 32.00 mm, whereas the model's predictions for these maximum values range from 24.42 mm to 39.40 mm.

The trained model is then utilized to predict rainfall values during the testing phase, and the time series plots between observed and predicted rainfall values for the four grids are presented in Figure 6. Compared to the training phase, the model performs well during the testing phase across all four grids, as evident in Figure 6. One of the reasons for this could be the rainfall values during the testing phase itself, as the maximum values are lower compared to those during the training phase. The actual maximum rainfall values range from 10.27 mm to 17.82 mm for the testing data. The model predictions for these maximum values range from 12.80 mm to 18.54 mm, with differences between the predicted and actual maximum values ranging from 0.01 mm to 6.18 mm.

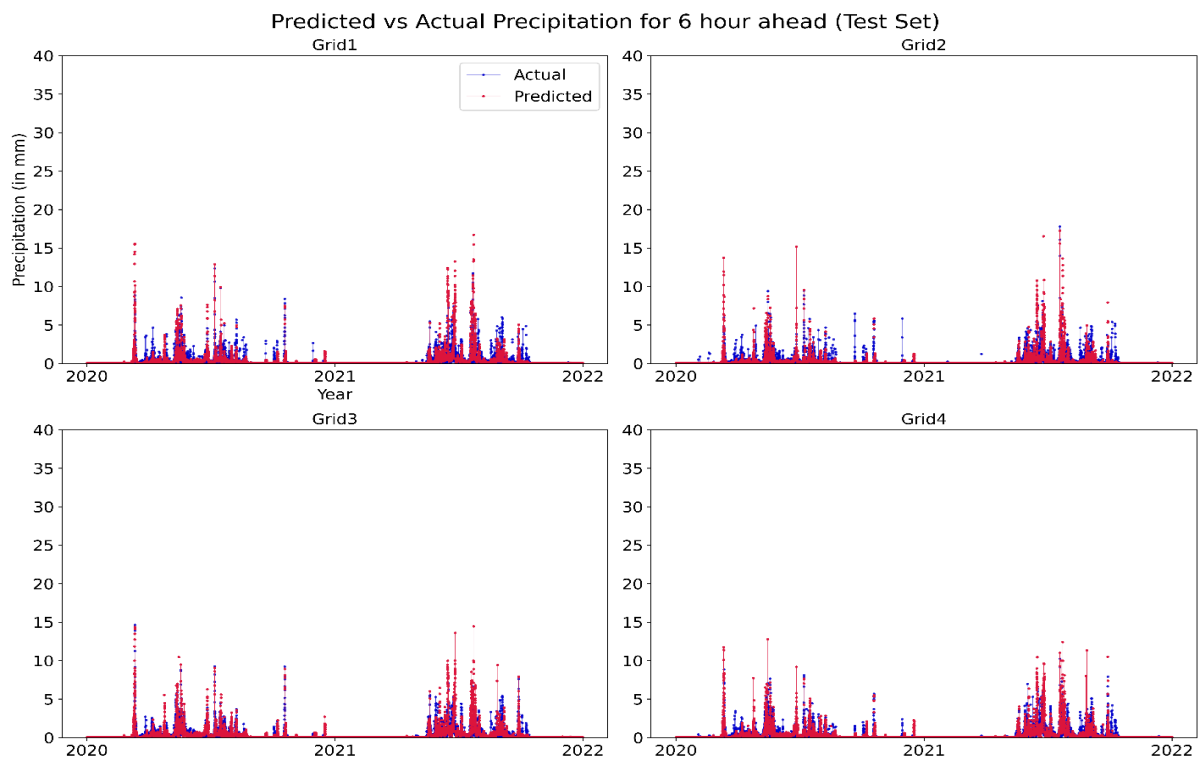

**Figure 6**. Time series comparisons of actual and forecasted rainfall across four grids at 6-hour intervals during the testing phase. The observed data points are represented by a blue color, whereas forecasted data points are illustrated using a red color.

The ConvLSTM2D model is then utilized to predict rainfall values at 12-hour intervals using 11 predictors during the training phase. Figure 7 displays the time series plot depicting observed and predicted values during the testing phase for visual comparison. The time series plots are presented for all four grids, with observed values shown in blue and predicted values in red. Similar to the 6-hour intervals, the time series plot obtained for 12-hour intervals also indicates that the predicted values are in good agreement with the observed values, albeit with some underestimations and overestimations.

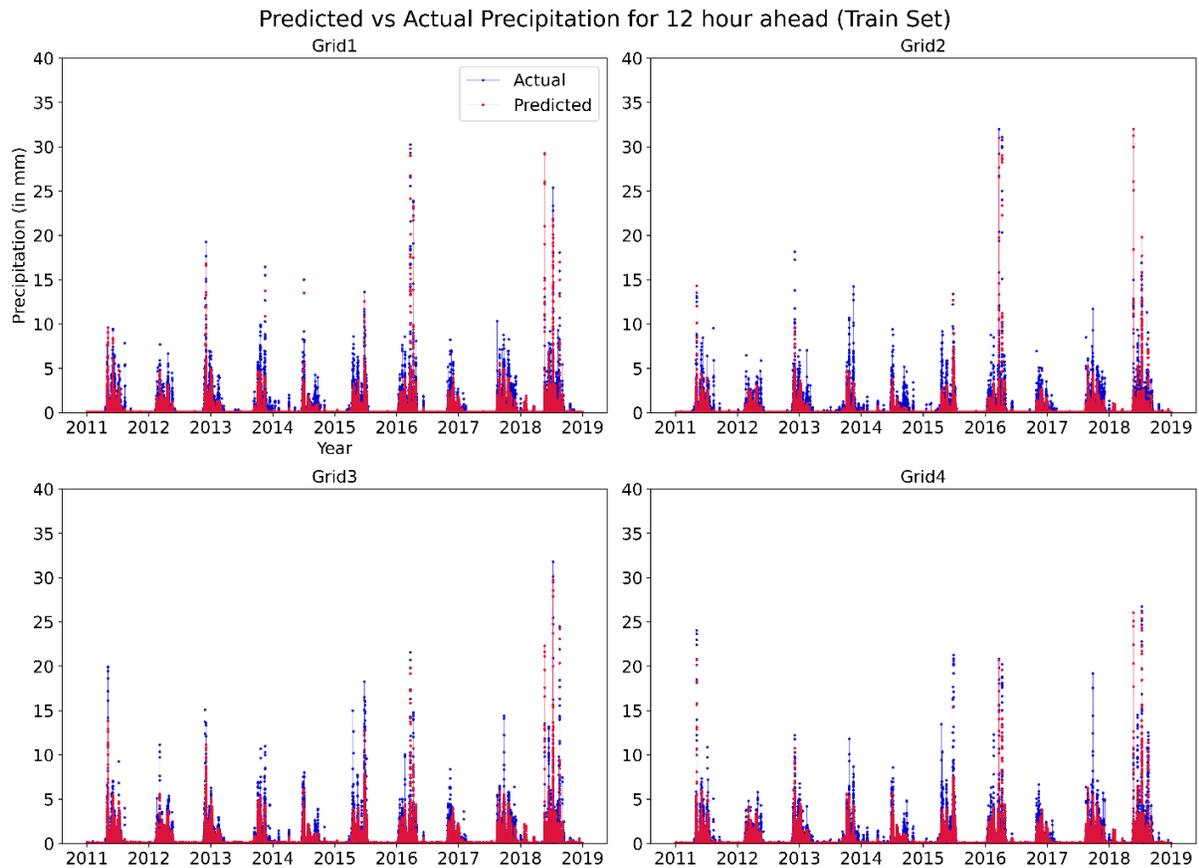

**Figure 7.** Time series comparisons of actual and forecasted rainfall across four grids at 12-hour intervals during the training phase. The observed data points are represented by a blue color, whereas forecasted data points are illustrated using a red color.

Figure 8 displays the time series plot between observed and predicted values during the testing phase at 12-hour intervals. The plot illustrates the model's ability to provide accurate predictions even at these intervals. Thus, visually, the ConvLSTM2D model developed in this study suggests that integrating correct atmospheric variables from a physics-based understanding of local climate processes with deep learning can enhance forecast skill in precipitation prediction. The results suggest that the ConvLSTM2D model is well-suited for spatiotemporal forecasting at finer scales, exhibiting commendable accuracy across different lead times.

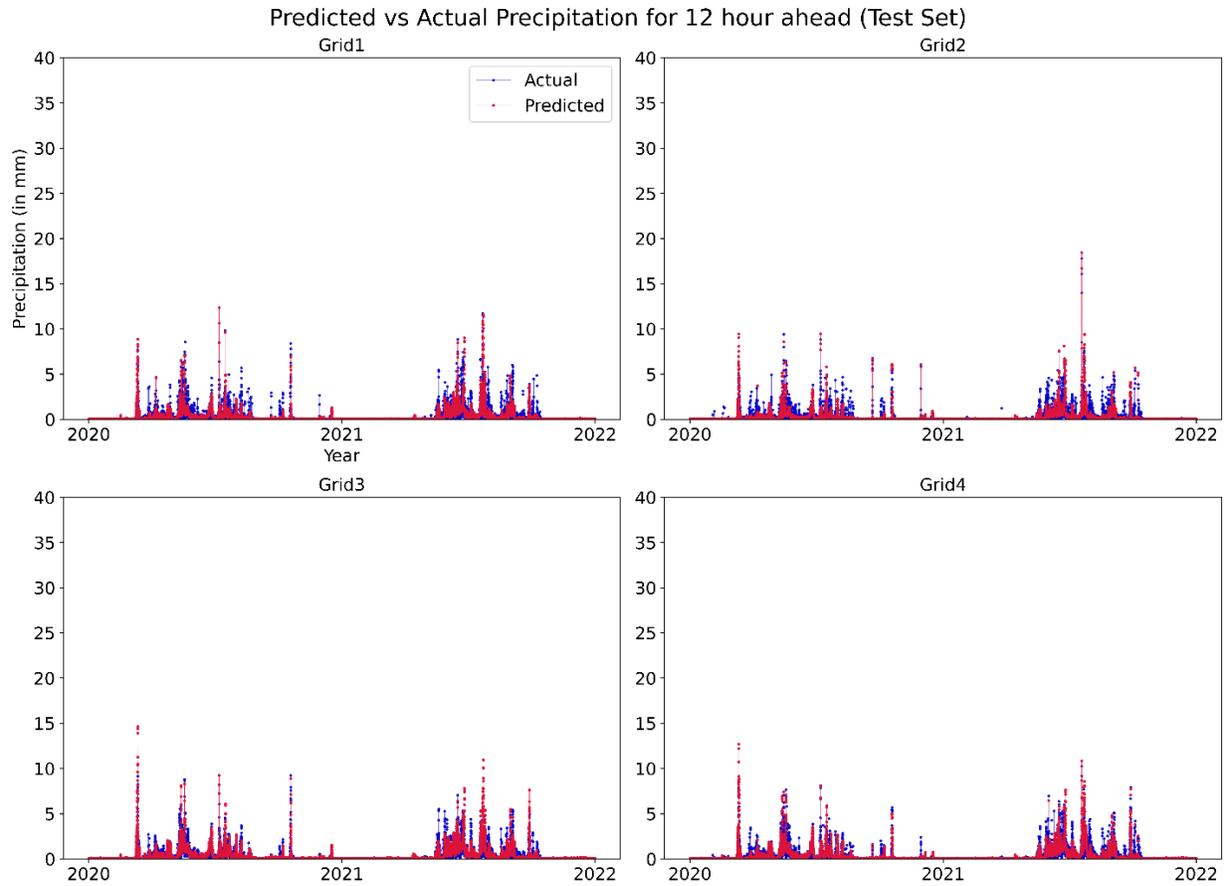

**Figure 8.** Time series comparisons of actual and forecasted rainfall across four grids at 6-hour intervals during the testing phase. The observed data points are represented by a blue color, whereas forecasted data points are illustrated using a red color.

### 4.3 Discussion

The ConvLSTM2D model has been utilized to predict rainfall at two time intervals: 6-hour and 12-hour intervals for both the training and testing phases. Although the model performs well for both intervals based on visual assessments from time series plots comparing observed and predicted values, it is essential to examine how the model's accuracy varies with different time steps. To achieve this, in addition to the time series plot, scatter plots can also be considered for visually comparing observed and predicted rainfall values. Figure 9 depicts the scatter plot illustrating the relationship between observed and predicted rainfall values during both the training and testing phases. The scatter plots for both 6-hour and 12-hour intervals are presented in one plot to enable a direct comparison of the model's accuracy across different time intervals. Moreover, the correlation values obtained between the observed and predicted rainfall values are also displayed in each subplot of the plots, denoted by 'R'. The scatter plots indicate strong alignment between predicted and observed rainfall across all four grids for both time intervals. There is consistent prediction of rainfall throughout the training phase, regardless of the time intervals considered. However, during training, the predicted rainfall from the 12-hour time interval shows greater conformity with observed rainfall compared to the 6-hour interval.

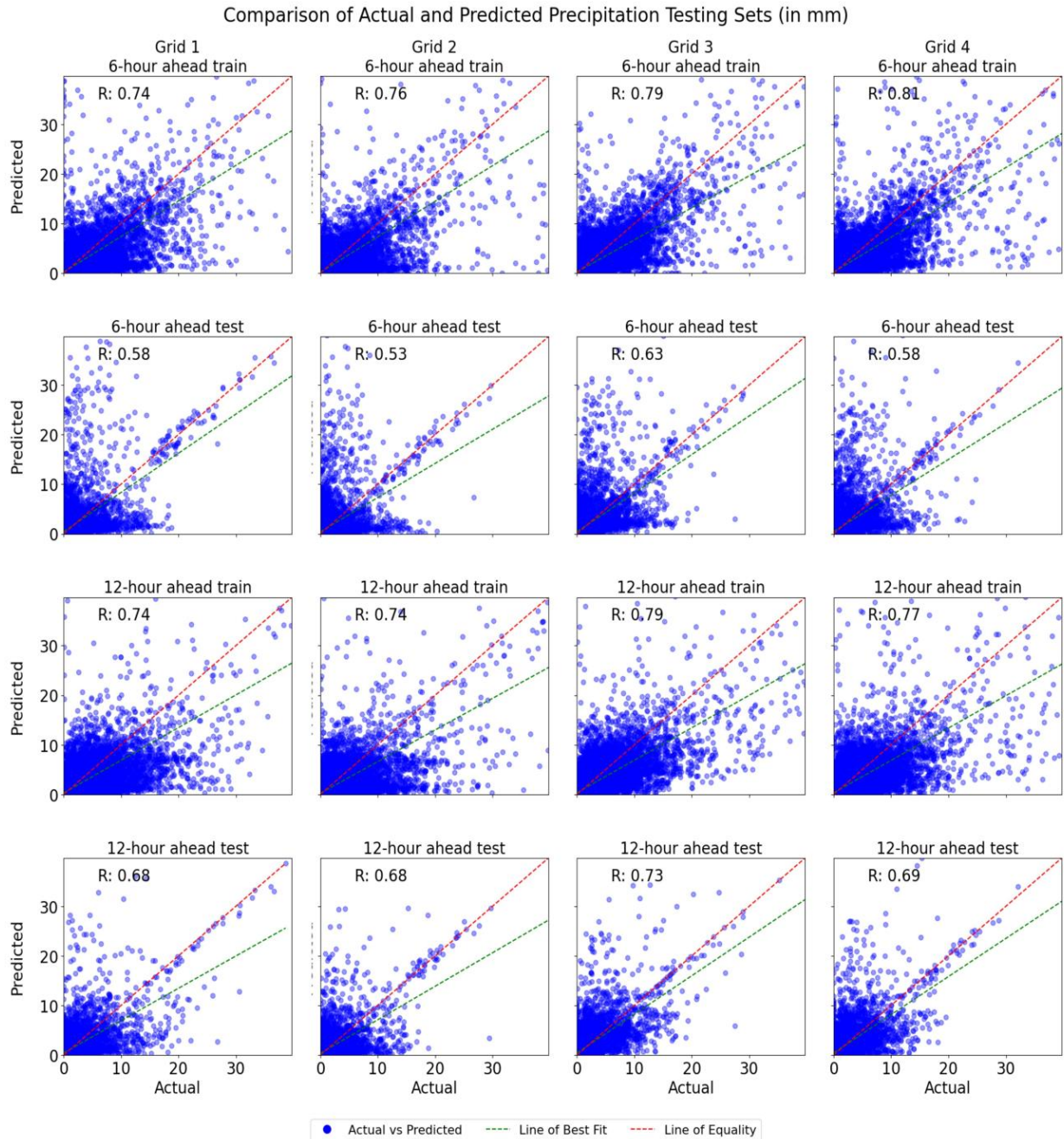

**Figure 9:** Scatter plots between observed and predicted rainfall values at 4 grids during testing and training phase. The first two rows are for 6-hour time intervals and the remaining two rows are for 12-hour time intervals.

The accuracy of the model in predicting rainfall at two time steps is also evaluated by calculating the values of the Correlation Coefficient (CC), Nash-Sutcliffe Efficiency (NSE), and Normalized

Root Mean Square Error (NRMSE). Table 2 presents the CC, NSE, and NRMSE values obtained during both the training and testing phases for all four grids for 6-hour and 12-hour time intervals. In the model training phase, correlation coefficients (CC) span between 0.74 and 0.81 for predictions made at 6-hour intervals, and 0.74 to 0.79 for 12-hour intervals, observed across all grids. During the testing phase, these CC values fluctuate between 0.53 and 0.63 for the 6-hour forecasts and 0.68 to 0.73 for the 12-hour forecasts, consistently across each grid. Hence, during the training phase, at grids 1 and 3, the predicted rainfall for both 6-hour and 12-hour intervals has the same CC value, whereas at grids 2 and 4, the predicted rainfall for 6-hour intervals has a higher CC than for 12-hour intervals.

The predictive accuracy is further assessed by NSE across all grids. NSE values indicate the model's ability to learn with high accuracy and capture variations. The NSE values for 6-hour intervals vary from 0.61 to 0.68 for training and from 0.42 to 0.51 for testing across all four grids. Similarly, the NSE values for 12-hour intervals range from 0.58 to 0.66 for training and from 0.47 to 0.58 for testing across the four grids. Therefore, during the testing phase, rainfall predicted at 12-hour intervals is better compared to 6-hour intervals. The range of NRMSE obtained during the training and testing phases for 6-hour intervals is from 1.47 to 2.12 and 1.95 to 2.14, respectively. For the 12-hour time intervals, the range is from 1.57 to 1.92 during training and 1.93 to 2.07 during testing.

**Table 2:** Values of CC, NSE, and NRMSE obtained during the training and testing phases for four grids for both 6-hour and 12-hour time intervals.

| Grids | | 6-hour Time Intervals | | | 12-hour Time Intervals | | |
|---|---|---|---|---|---|---|---|
| | | CC | NSE | NRMSE | CC | NSE | NRMSE |
| 1 | Training | 0.74 | 0.61 | 1.85 | 0.74 | 0.58 | 1.92 |
| | Testing | 0.58 | 0.45 | 2.12 | 0.68 | 0.48 | 2.07 |
| 2 | Training | 0.76 | 0.62 | 1.62 | 0.74 | 0.6 | 1.74 |
| | Testing | 0.53 | 0.42 | 2.14 | 0.68 | 0.47 | 1.98 |
| 3 | Training | 0.79 | 0.64 | 1.58 | 0.79 | 0.62 | 1.68 |
| | Testing | 0.63 | 0.51 | 1.95 | 0.73 | 0.58 | 1.93 |

| 4 | Training | 0.81 | 0.68 | 1.47 | 0.77 | 0.66 | 1.57 |
|---|----------|------|------|------|------|------|------|
|   | Testing  | 0.58 | 0.46 | 2.02 | 0.69 | 0.5  | 1.94 |

It is also important to mention at this point that previous research, which relied on numerical weather prediction models, encountered limitations in defining broad domains, initial conditions, and boundary conditions (Schultz et al., 2021; Hess and Boers, 2022). These constraints frequently created complexity and uncertainty in the forecasting process (Warner et al., 1997). Deep learning models, like the ConvLSTM2D model discussed in our study, overcome these constraints by effectively utilizing data without the need for extensive domain specifications or detailed boundary and initial conditions, thus offering a more direct and potentially more accurate method for localized weather event predictions.

One limitation of this study is that the scale of predictions should be narrowed down to lower spatial resolutions to make informed decisions at the urban scale. Therefore, the predictors and predictand variables should be available at a finer grid resolution, preferably below 0.25°. Future research should focus on improving precipitation prediction at a finer resolution with available information on the input variables at a coarser resolution. This can be accomplished by developing a physics-informed machine learning model that downscales precipitation using the input predictors. This could be a future scope of this research.

5.Conclusions

The present study represents an initial exploration of physics-informed deep learning-based rainfall prediction for fine spatial and temporal scales. The study utilized ConvLSTM2D, which is suitable for capturing spatial and temporal dimensions to learn and predict rainfall. Appropriate hyperparameters and model structures suitable for predicting rainfall with high accuracy were developed. It was found that physics-based variables associated with rainfall occurrence can be potential predictors as they represent the rainfall phenomena. Moreover, it was demonstrated that the input of such variables into the ConvLSTM2D model is capable of predicting rainfall.

Eleven predictor variables were considered to obtain the target variable (precipitation) across four grids in Mumbai. The predictors and target variables are ERA5 reanalysis products only. Hourly data from 2011 to 2022 were used, with 85% of the data utilized for training purposes and the remaining 25% for testing the model. Model accuracy during the training and testing phases was determined using CC, NSE, and NRMSE. The values of CC, NSE, and NRMSE obtained during both training and testing phases indicate the ConvLSTM2D model's proficiency in accurately predicting rainfall patterns across all four analyzed grids. Additionally, during the testing phase of the model, the predicted rainfall from 12-hour time intervals aligns well with the observed rainfall compared to the 6-hour time intervals.

Usually, the deep learning-based models learn from previous data, and ample data are sufficient to train the model with high prediction skills. However, in the case of the numerical model, initial and boundary conditions introduce uncertainties associated with prediction. Predicting rainfall in tropical regions is extremely difficult due to the complexity of understanding atmospheric behavior. In India, these challenges are amplified by the erratic nature of monsoon intra-seasonal oscillations, leading to significant fluctuations in rainfall over short durations. Forecasting extreme weather events, such as heavy precipitation, is essential to prevent human and economic losses. Accurate and timely warnings are powerful tools for capacity building in local disaster management and response agencies. We concluded that understanding the physical processes, atmospheric conditions, and associated variables is helpful in developing a well-performing rainfall prediction model. The model can be further integrated with city-level authorities and serve as a form of community-based weather forecast service to provide rainfall forecasts.

**CRediT authorship contribution statement**

Ajay Devda: Conceptualization, Methodology, Data Curation, Software, Investigation, Supervision, Writing – original draft

Akshay Sunil: Conceptualization, Methodology, Data curation, Supervision, Writing- Original draft preparation.

Murthy R: Data Curation, Software, Validation

B Deepthi: Data Curation, Software, Writing- Reviewing and Editing

**Code availability section**

The source codes for this project are available through the following link: https://github.com/akshaysunil172/Utilizing-Physics-Informed-ConvLSTM2D-Models-for-Finer-Spatial-and-Temporal-Resolution